\pdfoutput=1

\documentclass[11pt]{article}

\usepackage{acl}

\usepackage{times}
\usepackage{latexsym}

\usepackage[T1]{fontenc}

\usepackage[utf8]{inputenc}

\usepackage{times}
\usepackage{latexsym}
\usepackage{algorithm, algorithmic}
\usepackage{amsmath} 
\usepackage{amssymb}
\usepackage{multirow}
\usepackage{makecell}
\usepackage{hhline}
\usepackage{xspace}
\usepackage{booktabs}
\usepackage{float}
\usepackage{graphicx}

\usepackage{amssymb}
\usepackage{pifont}
\usepackage{bm}
\usepackage{enumitem}
\usepackage{diagbox}
\setlist[itemize]{leftmargin=*}
\setitemize[1]{itemsep=0pt,partopsep=0pt,parsep=\parskip,topsep=0pt}
\usepackage{microtype}

\title{Hero-Gang Neural Model For Named Entity Recognition}

\author{%
Jinpeng Hu$^{\heartsuit}$, \hspace{0.2cm}
Yaling Shen$^{\heartsuit}$, \hspace{0.2cm}
Yang Liu$^{\heartsuit}$ \hspace{0.2cm} \\
 \textbf{Xiang Wan}$^{\heartsuit\diamondsuit\dag}$, \hspace{0.2cm} \textbf{Tsung-Hui Chang}$^{\heartsuit\dag}$ \\
$^{\heartsuit}$Shenzhen Research Institute of Big Data, The Chinese University of Hong Kong, \\Shenzhen, Guangdong, China \hspace{0.2cm} \\
$^{\diamondsuit}$Pazhou Lab, Guangzhou, 510330, China \hspace{0.2cm} \\
\texttt{
\{jinpenghu, yalingshen, yangliu5\}@link.cuhk.edu.cn} \\
\texttt{wanxiang@sribd.cn} \hspace{0.2cm}
\texttt{changtsunghui@cuhk.edu.cn}
}

\begin{document}
\maketitle

\def\thefootnote{\dag}\footnotetext{Corresponding author.}
\renewcommand{\thefootnote}{\arabic{footnote}}

\begin{abstract}
Named entity recognition (NER) is a fundamental and important task in NLP, aiming at identifying named entities (NEs) from free text. Recently, since the multi-head attention mechanism applied in the Transformer model can effectively capture longer contextual information, Transformer-based models have become the mainstream methods and have achieved significant performance in this task. Unfortunately, although these models can capture effective global context information, they are still limited in the local feature and position information extraction, which is critical in NER. In this paper, to address this limitation, we propose a novel Hero-Gang Neural structure (HGN), including the Hero and Gang module, to leverage both global and local information to promote NER. Specifically, the Hero module is composed of a Transformer-based encoder to maintain the advantage of the self-attention mechanism, and the Gang module utilizes a multi-window recurrent module to extract local features and position information under the guidance of the Hero module. Afterward, the proposed multi-window attention effectively combines global information and multiple local features for predicting entity labels. Experimental results on several benchmark datasets demonstrate the effectiveness of our proposed model.\footnote{Our code is released at \url{https://github.com/jinpeng01/HGN}.}
\end{abstract}
\begin{figure*}[t]
\centering
\includegraphics[width=0.99\textwidth, trim=0 10 0 5]{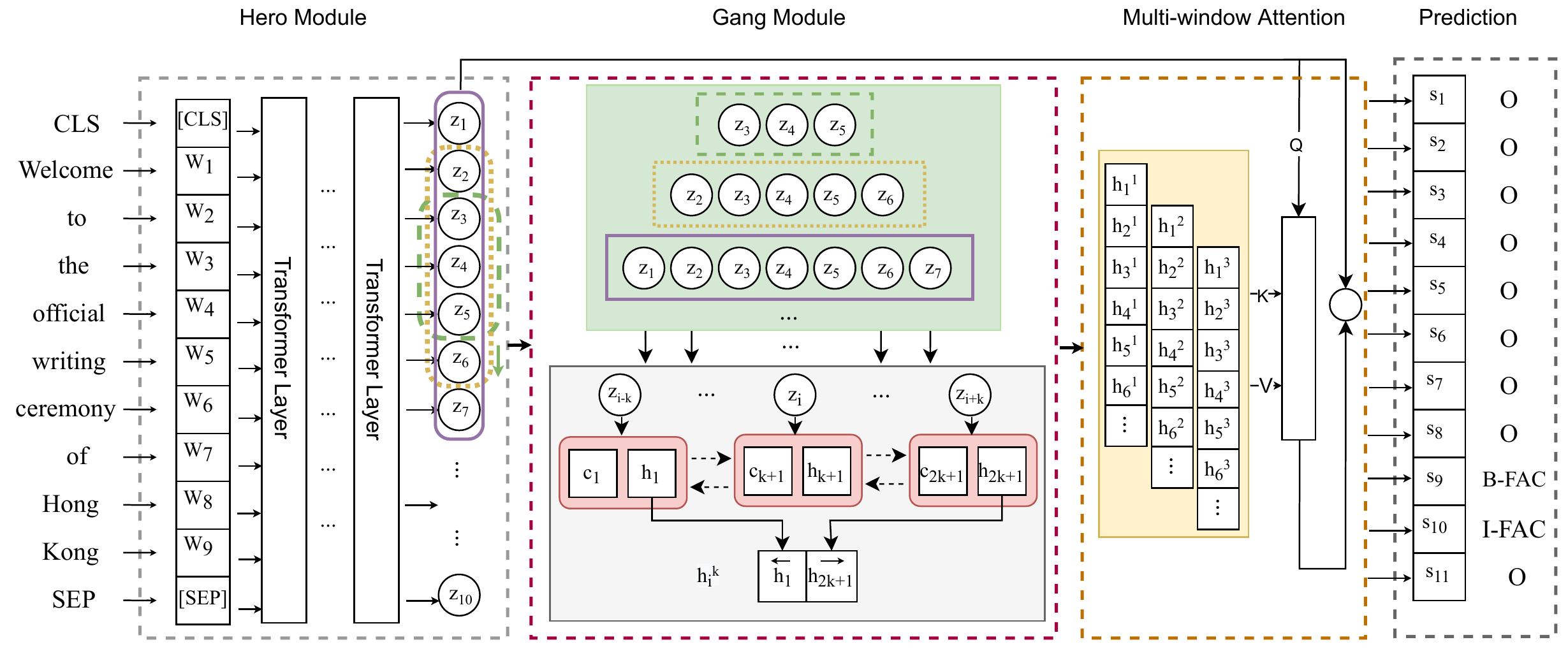}
\caption{The overall architecture of our proposed model. From left to right are the Hero module, Gang module, and multi-window attention, respectively, shown in different dashed boxes.
The purple solid frame, green, and yellow dashed frames in the Hero module are sliding windows with different window sizes.
The green box in the Gang module shows the multiple sub-sequences generated by the sliding windows for $\mathbf{z}_{4}$, and the grey box represents the bidirectional recurrent mechanism that is used to capture local features from these sub-sequences. 
Note that $\overleftarrow{\mathbf{h}_{1}}$ and $\overrightarrow{\mathbf{h}_{2k+1}}$ are the last hidden states of backward and forward recurrent structures.
The extracted local information is shown in the yellow box with its corresponding sub-sequences in the green box.
}
\label{fig:architecture}
 \vskip -1em
\end{figure*}

\section{Introduction}
Named entity recognition (NER) is one of the most important and fundamental research topics in natural language processing (NLP), which recognizes named entities (NEs), such as person, location, disease from raw text.
NER has attracted substantial attention in the past decades owing to its importance in downstream tasks, e.g., knowledge graph construction \cite{bosselut2019comet}, question-answering \cite{pergola2021boosting}, and relation extraction \cite{he2019classifying}.

In the early stage, the popular methods for solving NER are some traditional machine learning methods, e.g., Hidden Markov Model (HMM) \cite{morwal2012named} and conditional random field (CRF) \cite{mozharova2016combining}, which require extracting features manually, making the process inefficient and time-consuming.
With the breakthrough of recurrent neural networks (RNN) in NLP, Long short-term memory (LSTM) \cite{hochreiter1997long} and Gated Recurrent Unit (GRU) \cite{GRU} have become mainstream methods for this task and have achieved promising results since neural networks can automatically extract features from the sequence and also take each token's position information into consideration \cite{lample2016neural,chiu2016named,huang2015bidirectional}.
Nevertheless, RNN fails to perform well with long sequences due to the gradients exploding and vanishing.
In recent years, Transformer-based models \cite{vaswani2017attention} have become mainstream methods because these models are able to capture long-term dependencies with the help of multi-head attention and thus provide better global context information, especially for long sequences \cite{biobert,yangxlnet}.
However, these Transformer-based models usually are insensitive to the local context since the representation of each token is computed by the canonical point-wise dot-product self-attention \cite{li2019enhancing,huang2021missformer}.
Besides, although some studies \cite{shaw2018self,BERT,liu2019roberta} have been proposed to inject position information into Transformer, they are still inadequate to help Transformer obtain appropriate position information \cite{huang2020improve,qu2021explore}.
In other words, the self-attention mechanism is effective in overcoming the constraints of RNN from the perspective of long-sequence context information extraction, but is inferior to RNN in terms of local contextual and position information extraction.
Yet, both long-term dependencies and local context information are essential for the NER model to correctly identify entities.

Thus, to alleviate the shortcomings in RNN and Transformers while maintaining their respective strengths, in this paper, we propose a novel Hero-Gang Neural model to leverage both global and local contextual information to improve NER.
In doing so, on the one hand, we utilize a Transformer-based sequence encoder (i.e., Hero module) to extract effective global contextual information with the help of the self-attention mechanism.
On the other hand, a multi-window recurrent unit (i.e., Gang module) is applied to extract local features from multiple sub-sequences under the guidance of the extracted global information.
Afterward, we propose to use multi-window attention to elaborately combine global and local contextual features.
The performance of our proposed model significantly outperforms the strong baseline models on several NER benchmark datasets (including both general and biomedical domains) and achieves new state-of-the-art results on some datasets.

\section{Method}
\textcolor{black}{
NER is usually performed as a sequence labeling problem.
In detail, given a sequence of $X =x_{1},x_{2},...,x_{N}$ with $N$ tokens, we aim to learn a function that maps the input sequence into another one with the corresponding label $\hat{Y}=\hat{y_{1}},\hat{y_{2}},\hat{y_{3}},...,\hat{y_{n}}$ in the same length.
As summarized in Figure \ref{fig:architecture}, the Transformer-based models (e.g., BERT \cite{BERT}, XLNET \cite{yangxlnet}) are regarded as the Hero module to model the entire sentence for global sequence information extraction and the Gang module is responsible for local and relative position information extraction.
Afterward, we employ the multi-window attention to elaborately combine these different features (i.e., features extracted from the Hero and Gang modules), which is then used to predict labels for each token. Therefore, the aforementioned process can be formulated as:
\begin{equation}
\setlength\abovedisplayskip{6pt}
\setlength\belowdisplayskip{6pt}
    \hat{Y} = f(X, \text{H}(X), \text{G}(X)),
\end{equation}
where $\text{H}(\cdot)$ and $\text{G}(\cdot)$ refer to the Hero and Gang modules, respectively, and the details of them are presented in the following subsections.
}

\subsection{Hero Module}
\textcolor{black}{
\textcolor{black}{The role of the Hero module in our proposed model is similar to that of the leader in a team, who is responsible for providing guidance, offering instructions, giving directions, and assigning sub-tasks to fellow memberships. Therefore, the Hero module is required to have a comprehensive understanding of the task, including overall and local progress.}
Thanks to the characteristics of the multi-head self-attention mechanism, Transformer is powerful in modeling long sequences and can provide more effective global information than other counterpart models, and it has already achieved promising results in the NER task \cite{luo2020hierarchical,beltagy2019scibert}. 
Thus, we employ a Transformer-based encoder as our Hero module to obtain the global context information $\mathbf{z}_{i}$ for each token $x_{i}$ by
\begin{equation}
\setlength\abovedisplayskip{6pt}
\setlength\belowdisplayskip{6pt}
    [\mathbf{z}_{1},\mathbf{z}_{2},\cdots,\mathbf{z}_{N}] = f_\text{H} (x_{1},x_{2},...,x_{N}).
\end{equation}
Herein, $f_\text{H} (\cdot)$ refers to a pre-trained Transformer-based sequence encoder (e.g., BERT \cite{BERT} and BioBERT \cite{biobert}).
}
\textcolor{black}{The features $\mathbf{z}$ are then input to the Gang module for extracting local contextual features and their corresponding relative position information.}

\begin{table*}[t]
\footnotesize
\centering
\begin{tabular}{l|l|ccc|ccc|ccc}
\toprule[1pt]
\multicolumn{1}{l|}{\multirow{2}{*}{\textbf{Type}}} &\multicolumn{1}{l|}{\multirow{2}{*}{\textbf{Dataset}}} &\multicolumn{3}{c|}{\textsc{\textbf{Train}}}  & \multicolumn{3}{c|}{\textsc{\textbf{Val}}} & \multicolumn{3}{c}{\textsc{\textbf{Test}}}\\
\cmidrule(l){3-5}
\cmidrule(l){6-8}
\cmidrule(l){9-11}
\multicolumn{1}{c|}{}&\multicolumn{1}{c|}{}  & \textsc{\textbf{\#Sent.}}  & \textsc{\textbf{\#Ent.}}   & \textsc{\textbf{\#AS.}}   & \textsc{\textbf{\#Sent.}}  & \textsc{\textbf{\#Ent.}}   & \textsc{\textbf{\#AS.}}& \textsc{\textbf{\#Sent.}}  & \textsc{\textbf{\#Ent.}}   & \textsc{\textbf{\#AS.}}  \\ 
\midrule
\multirow{3}{*} {\makecell*[l]{\textsc{General}}}
& \textsc{{W16}} & {2.4k} & {1.5k} & {19.41}&  {1.0k}& {0.7k}& {16.26} &{3.9k}& {3.5k}& {16.08}\\
& \textsc{{W17}} & {3.4k} & {2.0k} & {18.48}&  {1.0k}& {0.8k}& {15.59} &{1.3k}& {1.1k}& {18.18}\\
& \textsc{{ON5e}} &  {59.9k} & {81.8k} & {18.17}&  {8.5k}& {11.1k}& {17.32} &{8.3k}& {11.3k}& {18.49}\\
\midrule
\multirow{3}{*} {\makecell*[l]{\textsc{Biomed}}}
&\textsc{{BC5-D}} &  {4.6k} & {4.2k} & {25.79}&  {4.6k}& {4.2k}& {25.52} &{4.8k}& {4.4k}& {25.92}\\
&\textsc{{BC2GM}} &  {12.6k} & {15.2k} & {28.14}&  {2.5k}& {3.0k}& {28.07} &{5.0k}& {6.3k}& {28.33}\\
&\textsc{{BC5-C}} &  {4.6k} & {4.2k} & {25.79}&  {4.6k}& {4.2k}& {25.52} &{4.8k}& {4.4k}& {25.92}\\
\bottomrule
 \end{tabular}
  \caption{The statistics of the six benchmark datasets w.r.t. their training, validation and test sets, including the number of sentences (\#Sent.), the number of entities (\#Ent.), and the averaged word-based length (\#AS.).}%
  \label{Tab:NERdataset}
\vspace{-3mm}
\end{table*}
\subsection{Gang Module}
\textcolor{black}{As introduced in the previous section, although pre-trained models are able to provide effective global contextual representation, it lacks the ability to extract local features and relative position information.
\textcolor{black}{Thus, we propose a multi-window recurrent module, named Gang, to enhance local information extraction.}
Recurrent structures (RS), such as LSTM, GRU, and tradition RNN are effective in extracting both local and relative position information from the sequence, owing to characteristics of the recurrent mechanism.
To better emphasize the local features of each word without being disturbed by long-distance information, we construct a sliding window with a fixed length to generate shorter sub-sequences, where each sub-sequence includes several consecutive elements in $\mathbf{z}$.
An additional advantage of this operation is that, in comparison with the whole sequence, the sub-sequence is much shorter so that it is easier to be modeled by the RS.
}

\textcolor{black}{
In detail, for a single sliding window with length $k$, each hidden state $z_{i}$ from the Hero module, the corresponding sub-sequence is $\mathbf{z}_{i-k},\mathbf{z}_{i-k+1},...,\mathbf{z}_{i},...,\mathbf{z}_{i+k-1},\mathbf{z}_{i+k}$ that includes $2k+1$ consecutive tokens.
This sub-sequence of length $2k+1$ contains rich local contextual information of $x_{i}$, and thus we utilize an RS to encode it for obtaining local semantic and relative position information.
To extract the local information of two directions, we utilize a bidirectional structure to encode this sequence span, where the forward RS computes a representation $\overrightarrow{\mathbf{h}_{2k+1}}$ from left to right, and the other backward RS computes a vector $\overleftarrow{\mathbf{h}_{1}}$ for the same sub-sequence in reverse.}
\textcolor{black}{We concatenate the $\overleftarrow{\mathbf{h}_{1}}$ and $\overrightarrow{\mathbf{h}_{2k+1}}$ as the local feature $\mathbf{h}_{i} = [\overleftarrow{\mathbf{h}_{1}}, \overrightarrow{\mathbf{h}_{2k+1}} ]$ for token $x_{i}$, and then we can obtain local features for each token in sequence $X$ via the similar way, denoted as $\mathbf{h}=\mathbf{h}_{1},\mathbf{h}_{2},\cdots,\mathbf{h}_{N}$.}

\textcolor{black}{In practice, we need to consider two situations.
First, each token might have multiple levels of local information, such as phrase-level and clause-level, which may affect the understanding of the current token.
Second, since different tokens or the same token in various contexts might have different relationships with their surrounding words, we need to consider more sub-sequences with varying lengths for obtaining more comprehensive local contextual information.
Therefore, we propose to utilize multiple sliding windows with different window sizes to extract richer local features to alleviate the above issues.
We assume that local features $\mathbf{h}^{1}, \mathbf{h}^{2}, \cdots, \mathbf{h}^{M}$ are extracted from different groups of sub-sequences, whose corresponding window lengths are $k^{1}, k^{2}, \cdots, k^{M}$. This process can be formulated as:
}
\textcolor{black}{
\begin{equation}
\setlength\abovedisplayskip{6pt}
\setlength\belowdisplayskip{6pt}
   \mathbf{h}^{1}, \mathbf{h}^{2}, \cdots, \mathbf{h}^{M} = \text{Gang} (k^{1}, k^{2}, \cdots, k^{M},\mathbf{z}),
\end{equation}
\textcolor{black}{
where $M$ is the number of sliding windows and $\mathbf{h}^{j}$ is a group of local features extracted from the corresponding sliding window with length $k^{j}$. The process is similar to the task assignment in the team, where different members are responsible for their own sub-tasks.
}}

\subsection{Multi-window Attention}
\textcolor{black}{We obtain global representation $\mathbf{z}$ from the Hero module and multiple local features $\mathbf{h}^{1}, \mathbf{h}^{2}, \cdots, \mathbf{h}^{M}$ from the Gang module.
Next, we apply the multi-window attention to effectively combine global contextual information and local features.
In doing so, two types of attention methods are proposed in our model: MLP-Attention and DOT-Attention, respectively.}
\vskip 0.2em
\noindent\textbf{MLP-Attention}~
\textcolor{black}{We first concatenate these local features with global information and obtain the intermediate state $\mathbf{m}$ by a fully connected layer.}
\begin{equation}
\setlength\abovedisplayskip{6pt}
\setlength\belowdisplayskip{6pt}
  \mathbf{m} = \text{MLP}([\mathbf{z},\mathbf{H}]),
\end{equation}
\textcolor{black}{where $\mathbf{H}=[\mathbf{h}^{1},\mathbf{h}^{2}, \cdots, \mathbf{h}^{M}]$ and $\mathbf{m}$ have the same dimension as $\mathbf{z}$. MLP represents a fully connected layer.
Then $\mathbf{m}$ is used as a query vector and $[\mathbf{z},\mathbf{H}]$ serves as the key and value matrix.}
\textcolor{black}{The final token representation can be computed by}
\begin{equation}
\setlength\abovedisplayskip{6pt}
\setlength\belowdisplayskip{6pt}
  \mathbf{s} = \text{softmax}(\mathbf{m}([\mathbf{z},\mathbf{H}])^{\top})[\mathbf{z},\mathbf{H}].
\end{equation}
\vskip 0.2em
\noindent\textbf{DOT-Attention}~
\textcolor{black}{Instead of using a fully connected layer to generate a query vector, in this approach, we directly regard $\mathbf{z}$ as the query vector and $\mathbf{H}$ as the key and value matrix.
We can obtain the final local feature by}
\begin{equation}
\setlength\abovedisplayskip{6pt}
\setlength\belowdisplayskip{6pt}
  \mathbf{u} = \text{softmax}(\mathbf{z}(\mathbf{H})^{\top})\mathbf{H}.
\end{equation}
\textcolor{black}{
Since $\mathbf{u}$ is a weighted sum of different local features without considering global information, we use the sum of $\mathbf{u}_{i}$ and $\mathbf{z}_{i}$ as the final representation for each token $x_{i}$.
Thus, the final representation can be obtained by}
\begin{equation}
\setlength\abovedisplayskip{6pt}
\setlength\belowdisplayskip{6pt}
  \mathbf{s} = \{\mathbf{z}_{1}+\mathbf{u}_{1},\mathbf{z}_{2}+\mathbf{u}_{2}, \cdots, \mathbf{z}_{N}+\mathbf{u}_{N}\}.
\end{equation}

\textcolor{black}{After obtaining the final representation from MLP-Attention or DOT-Attention, $\mathbf{s}$ is sent to the corresponding classifier implemented by the softmax function to predict the distribution of labels for each token in $X$.}

\begin{table*}[t]
\footnotesize
\centering
\resizebox{.99\textwidth}{!}{
\begin{tabular}{lccccccccc}
\toprule[1pt]
\multicolumn{1}{l}{\multirow{2}{*}{\textbf{Methods}}} &\multicolumn{3}{c}{\textsc{\textbf{W16}}}  & \multicolumn{3}{c}{\textsc{\textbf{W17}}} & \multicolumn{3}{c}{\textsc{\textbf{ON5e}}}\\ 
\cmidrule(l){2-4}
\cmidrule(l){5-7}
\cmidrule(l){8-10}
\multicolumn{1}{c}{}  & \textsc{\textbf{P}}        & \textsc{\textbf{R}}   & \textsc{\textbf{F-1}}& \textsc{\textbf{P}}     & \textsc{\textbf{R}} & \textsc{\textbf{F-1}}  & \textsc{\textbf{P}} & \textsc{\textbf{R}} & \textsc{\textbf{F-1}}\\ 
\cmidrule(l){1-10}
\textbf{with incorporating extra resources} & &  & &  & &  &  & & \\
\textsc{SANER} \cite{nie2020named} & {-} & {51.27} & {55.01}&  {-}& {{49.45}}& {50.36} & {-} & {-} & {-} \\
\textsc{AESUBER} \cite{AESUBER} & {-} & {-} & {55.14}&  {-}& {-}& {50.68} & {-} & {-} & {90.32} \\
\textsc{Hire-NER} \cite{luo2020hierarchical} & {-} & {-} & {-}&  {-}& {-}& {-} & {-} & {-} & {90.30} \\
\textsc{CL-KL} \cite{wang2021improving} & {-} & {-} & \textbf{58.98}&  {-}& {-}& \textbf{60.45} & {-} & {-} & {-} \\
\textsc{Syn-LSTM-CRF} \cite{xu2021better} & {-} & {-} & {-}&  {-}& {-}& {-} & {90.14} & {91.58} & \textbf{90.85} \\

\cmidrule(l){1-10}
\textbf{without extra resources} & &  & &  & &  &  & & \\
\textsc{CNN-BiLSTM-CRF} \cite{chiu2016named} & {-} & {-} & {-}&  {-}& {-}& {-} & {86.04} & {86.53} & {86.28} \\
\textsc{BERT} \cite{BERT} & {-} & {49.02} & {54.36}&  {-}& {46.73}& {49.52} & {-} & {-} & {89.16} \\
\textsc{XLNET} \cite{yangxlnet}& {55.94} & {57.46} & \textbf{56.69}& {58.68} & {49.18}& \textbf{53.51}& {89.72}& {91.05}& \textbf{90.38} \\
\textsc{ASTRA} \cite{wang2020astral} & {-} & {-} & {-}& {-} & {-}& {49.72}& {-}& {-}& {89.44} \\
\textsc{BARTNER} \cite{yan2021unified} & {-} & {-} & {-}& {-} & {-}& {-}& {89.99}& {90.77}& \textbf{90.38} \\

\cmidrule(l){1-10}
\textsc{HGN (BERT) (concat)} & {56.06} & {55.61}   & {55.84}
& {57.41}  & {45.45} & {50.74}  
& {89.20}& {89.85}& {89.52}\\
\textsc{HGN (BERT) (add)} & {54.63} & {55.38}   & {55.01}
&  {58.46}  & {45.55} & {51.20} 
& {89.16}& {90.01}& {89.58}\\
\textsc{HGN (BERT) (mlp)} &  {57.72} & {55.66}   & {56.67}
& {59.26}  &  {50.70} &  {54.65}  
& {89.19}&  {90.24}&  {89.71}\\
\textsc{HGN (BERT) (dot)} & {57.51} &  {56.00}   &  {56.75}
& {60.09}  & {48.29} & {53.55}  
&  {89.32}& {90.11}&  {89.71}\\

\textsc{HGN (XLNET) (concat)} & {57.48} & {57.90}   & {57.69} 
& {63.39} & {49.27} & {55.45} 
& {89.92}& {91.35}& {90.63}\\
\textsc{HGN (XLNET) (add)} & {57.31} & {58.05}   & {57.68} 
& {59.11} & {48.36} & {53.20} 
& {90.10}& {91.39}& {90.74}\\
\textsc{HGN (XLNET) (mlp)} 
& {58.91} & {59.89}   & {59.39}
& {63.16} &  {52.27} & {57.20} 
& {90.29}& {91.56}& \textbf{90.92}\\
\textsc{HGN (XLNET) (dot)} 
& {59.74} &  {59.26}   & \textbf{59.50}
& {62.49} & {53.10} & \textbf{57.41} 
& {90.10}& {91.64}& {90.86}\\
\bottomrule
 \end{tabular}}
   \caption{Comparisons of our proposed models with previous studies on the W16, W17, and ON5e, respectively, with respect to precision, recall, and F-1 score for NER. 
  Previous studies are divided into two parts from top to bottom, representing methods requiring extra resources and without such requirements, respectively.
   }%
  \label{Tab:NER}
\vspace{-2mm}
\end{table*}

\section{Experiments Settings}
\subsection{Dataset and Metrics}
\textcolor{black}{
In our experiments, six datasets are used in our experiments, WNUT17 (W17) \cite{w16}, WNUT16 (W16) \cite{w17}, OntoNotes 5.0 (ON5e) \cite{pradhan2013towards}, BC5CDR-disease (BC5-D), BC2GM, and BC5CDR-chem (BC5-C).
The W17 and W16 are social media benchmark datasets constructed from Twitter, and ON5e is a general domain dataset consisting of diverse sources like telephone conversations, newswire, etc.
\textcolor{black}{BC5CDR, including both BC5-D and BC5-C, is a dataset used for the BioCreative V Chemical Disease Relation Task and contains chemical and disease mentions, where humans manually annotate the annotations.} 
BC2GM is the dataset that is usually utilized for the BioCreative II gene mention tagging task and contains 20000 sentences from the abstracts of biomedical publications.
For all datasets, we utilize the official splits for a fair evaluation and the statistics of the datasets are shown in Table \ref{Tab:NERdataset}.
Besides, we follow previous studies that the final models are trained on training and validation sets on each dataset except the ON5e dataset.
}

\textcolor{black}{For metrics, we exploit the same evaluation metrics used by previous works where precision (P), recall (R), and F-1 score are reported to evaluate the performance of our model.
%
}

\subsection{Implementation Details}
\textcolor{black}{
We implement our model based on transformers \cite{wolf2020transformers}\footnote{\url{https://github.com/huggingface/transformers}} and employ pre-trained models to obtain global contextualized representation.
Specifically, for general domain datasets (i.e., W16, W17 and ON5e), we use BERT-cased-large \cite{BERT}\footnote{We obtain the pre-trained BERT from \url{https://github.com/google-research/bert}.} and XLNET-large-cased \cite{yangxlnet}\footnote{We obtain XLNET from \url{https://github.com/zihangdai/xlnet}.} as our Hero module.
For biomedical datasets, BioBERT \cite{biobert}\footnote{We obtain BioBERT from \url{https://github.com/dmis-lab/biobert}} is utilized to obtain global information.
}
\textcolor{black}{
We follow their default settings for all BERT, XLNET, and BioBERT: 24 layers of self-attention with 1024 dimensional embeddings.
For hyperparameters of the Gang module, the hidden sizes of bidirectional recurrent structures for each window size are half of the embedding dimension from the output of the Hero module (i.e., 512).
During the training process, we use Adam \cite{adam} to optimize the negative log-likelihood loss function.
More training details are shown in the Appendix \ref{appendix:hyperparameter}.
Besides, we also compare four operations to combine different level features from the Hero and Gang module: MLP-Attention, DOT-Attention, concatenation, and summation, respectively, where concatenation is to connect all features directly through $\mathbf{s} = [\mathbf{h}^{1}, \mathbf{h}^{2},\cdots, \mathbf{h}^{M},\mathbf{z}]$,  and summation is to add up these features by $\mathbf{s} = \mathbf{h}^{1}+\mathbf{h}^{2}+\cdots+\mathbf{h}^{M}+\mathbf{z}$.
}

\subsection{Baselines}
To explore the impact of our proposed model, we compare our model to the previous studies.
For general domain, following baselines are compared in our experiment on W16, W17 and ON5e.
\begin{itemize}
    \item \textbf{\textsc{CNN-BiLSTM-CRF}} \cite{chiu2016named} utilizes a hybrid bidirectional and CNN architecture to detect word-and character-level features.
    \item \textbf{\textsc{BERT}} \cite{BERT} is a pre-trained language model and we apply it to the NER task by direct fine-tuning.
    \item \textbf{SANER} \cite{nie2020named}, \textbf{CL-KL} \cite{wang2021improving} and \textbf{\textsc{AESUBER}} \cite{AESUBER} improve entity recognition by leveraging syntactic information or semantically relevant texts.
    \item \textbf{\textsc{Hire-NER}} \cite{luo2020hierarchical} utilizes both sentence-level and document-level representations to improve sequence labeling.
    \item \textbf{\textsc{Syn-LSTM-CRF}} \cite{xu2021better} integrates the structured information by graph-encoded representations obtained from GNNs.
    \item \textbf{\textsc{BARTNER}} \cite{yan2021unified} formulates NER tasks as a span sequence generation problem.
\end{itemize}
 In addition, we also compare our proposed model with the following baselines on the aforementioned biomedical datasets:
\begin{itemize}
    \item \textbf{\textsc{MTM-CW}} \cite{wang2019cross}, \textbf{\textsc{BiLM}} \cite{sachan2018effective}, \textbf{\textsc{NCBI\_BERT}} \cite{peng2019transfer}, \textbf{\textsc{MT-BioNER}} \cite{2021A} utilize multi-task learning or transfer learning to enhance biomedical NER.
    \item \textbf{\textsc{BioBERT}} \cite{biobert} is a pre-trained model trained with a large amount of biomedical corpus and then applied by directly fine-tuning.
    \item \textbf{\textsc{\textsc{KeBio-LM}}} \cite{yuan2021improving} proposes a  biomedical pre-trained language model that incorporates knowledge from the Unified Medical Language System (UMLS).
\end{itemize}
Note that in both general and biomedical domains, our model \textbf{does not require external resources.}

\begin{table*}[t]
\footnotesize
\centering
\begin{tabular}{lccccccccc}
\toprule[1pt]
\multicolumn{1}{l}{\multirow{2}{*}{\textbf{Methods}}} &\multicolumn{3}{c}{\textsc{\textbf{BC5-D}}}  & \multicolumn{3}{c}{\textsc{\textbf{BC2GM}}} & \multicolumn{3}{c}{\textsc{\textbf{BC5-C}}}\\ 
\cmidrule(l){2-4}
\cmidrule(l){5-7}
\cmidrule(l){8-10}
\multicolumn{1}{c}{}  & \textsc{\textbf{P}}        & \textsc{\textbf{R}}   & \textsc{\textbf{F-1}}& \textsc{\textbf{P}}     & \textsc{\textbf{R}} & \textsc{\textbf{F-1}}  
& \textsc{\textbf{P}} & \textsc{\textbf{R}} & \textsc{\textbf{F-1}}\\ 
\cmidrule(l){1-10}
\textbf{with incorporating extra resources} & &  & &  & &  &  & & \\
\textsc{BiLM} \cite{sachan2018effective} & {-} & {-} & {-}& {81.81} & {81.57}& {81.69}& {-} & {-} & {-}\\
\textsc{MTM-CW} \cite{wang2019cross} & {-} & {-} & {-}& {82.10} & {79.42}& {80.74}& {-} & {-} & {-}\\


\textsc{KeBio-LM} \cite{yuan2021improving} & {-} & {-} & {86.10}& {-} & {-}& {85.10}& {-} & {-} & {93.30}\\
\textsc{MT-BioNER} \cite{2021A} & {-} & {-} & {-}& {84.42} & {85.14}& \textbf{84.78}& {93.29} & {94.69} & \textbf{93.98}\\
\cmidrule(l){1-10}

\textbf{without extra resources} & &  & &  & &  &  & & \\
\textsc{NCBI\_BERT} \cite{peng2019transfer} & {-} & {-} & {86.60}& {-} & {-}& {-}& {-} & {-} & \textbf{93.50}\\
\textsc{BioBERT} \cite{biobert} & {86.47} & {87.84} & \textbf{87.15}& {84.32} & {85.12}& \textbf{84.72}& {93.68}& {93.26}& {93.47}\\
\cmidrule(l){1-10}
\textsc{HGN (BioBERT) (concat)} & {85.90} & {88.81}   & {87.33} & 
{83.91} & {86.36} & {85.12} & 
{94.30} & {93.93} & {94.11} \\

\textsc{HGN (BioBERT) (add)} & {85.89} & {88.74}   & {87.29} & 
{85.21} & {85.50} & {85.35} & 
{94.01} & {94.57} & {94.29} \\

\textsc{HGN (BioBERT) (mlp)} & {86.70} & {88.86}   & {87.77} & 
{84.93} & {86.37} & \textbf{85.65} & 
{94.23} & {94.63} & {94.43} \\

\textsc{HGN (BioBERT) (dot)} & {86.27} & {89.51}   & \textbf{87.86} & 
{85.21} & {85.88} & {85.54} & 
{94.45} & {94.73} & \textbf{94.59} \\
\bottomrule
 \end{tabular}
   \caption{Comparisons of our proposed models with previous studies on the BC5-D, BC2GM, and BC5-C, respectively, for biomedical NER in iterms of precision, recall, and F-1 score. 
  Previous works are divided into two sections, indicating methods requiring extra resources and without such requirements.
   }%
  \label{Tab:BIO}
\vspace{-2mm}
\end{table*}

\section{Results and Analyses}

\subsection{General Domain NER}
\textcolor{black}{
In this subsection, to explore the effectiveness of our proposed model, we conduct experiments to compare our model with existing studies, and the results are reported in Table \ref{Tab:NER}.
There are several observations drawn from different aspects.
First, when we make a fair comparison without extra resources (e.g., BERT, XLNET, and ASTRA), our model obtains significant improvements on all datasets in terms of Precision, Recall, and F-1, which confirms the effectiveness of our proposed Hero-Gang neural structure. This is because multiple-level features can be reasonably encoded into the model and thus alleviate the limitations of Transformer in local feature extraction.
Second, although some complicated models enhance NER by incorporating extra knowledge, e.g., SANER uses augmented semantic information, Hire-NER utilizes two-level hierarchical contextualized representations, and CL-KL selects a set of semantically relevant texts to improve NER, our model achieves competitive results without such requirements.
This is because each word in the natural text usually has a closer relationship with its surrounding words, especially the adjacent words, such that features extracted by the Gang module can provide more valuable information for NER, and thus our model achieves promising performance.
Third, the XLNET-based model obtains better results than the BERT-based model, which indicates that XLNET can generate more effective representations on the NER task. The reason behind this might be that XLNET combines the permutation operation with the autoregressive technology to further improve representation learning, so that XLNET can provide a better text understanding than BERT.
}

\subsection{Biomedical NER}
We also compare our model with state-of-the-art models in the biomedical NER on the aforementioned datasets with all results reported in Table \ref{Tab:BIO}.
There are several observations.
First, we can see that our model outperforms existing methods, regardless of whether they introduce external knowledge, which further confirms the validity of our innovation in combining local and global features to enhance feature extraction.
Second, although some models utilize higher-level features, e.g., BIOKMNER leverages POS labels, syntactic constituents, dependency relations, and MTM-CW employs multi-task learning to train the model, our model can achieve better results through a simple Hero-Gang structure.
This means that local features extracted from the Gang module under the guidance of global information are also effective in assisting biomedical text representations and even show more significant potential than those special designs for the medical domain (i.e., domain-related multi-task learning).
Third, the models using the multi-window attention (i.e., DOT-Attention and MLP-Attention) outperform those using concatenation or summation.
This observation suggests that multi-window attention can elaborately weigh local features from different sliding windows to enhance feature combinations.

\subsection{Analyses}
\begin{figure}[t]
\centering
\includegraphics[width=0.49\textwidth, trim=0 10 0 10]{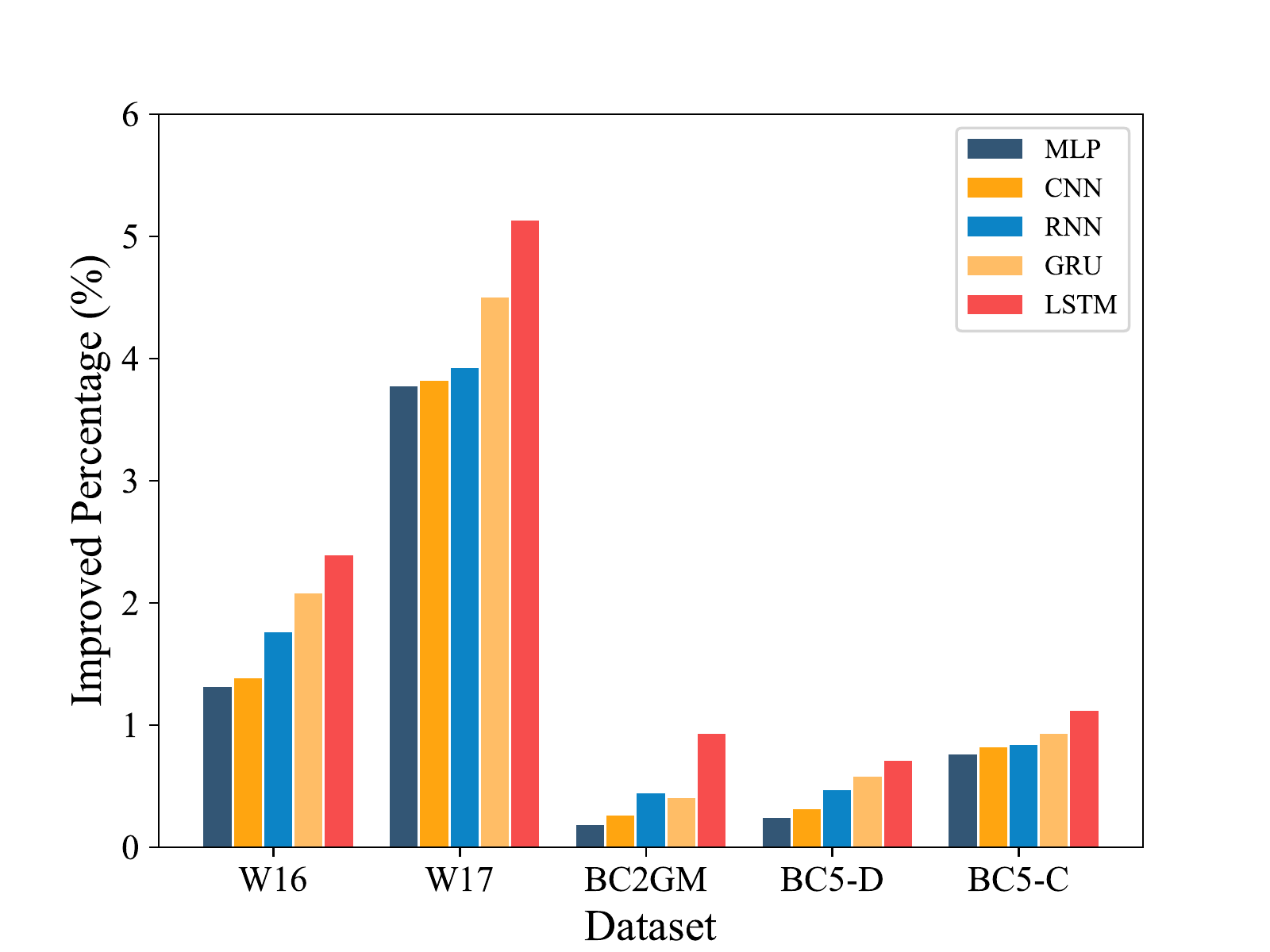}
\caption{\textcolor{black}{The improvement values (\%) compared to Base models (i.e., BERT for general domain datasets and BioBERT for biomedical datasets) in terms of F-1 score from different Gang modules, MLP, CNN, RNN, GRU, and LSTM, respectively.}}
\label{fig:position}
 \vskip -1em
\end{figure}

\begin{figure}[t]
\centering
\includegraphics[width=0.49\textwidth, trim=0 10 0 10]{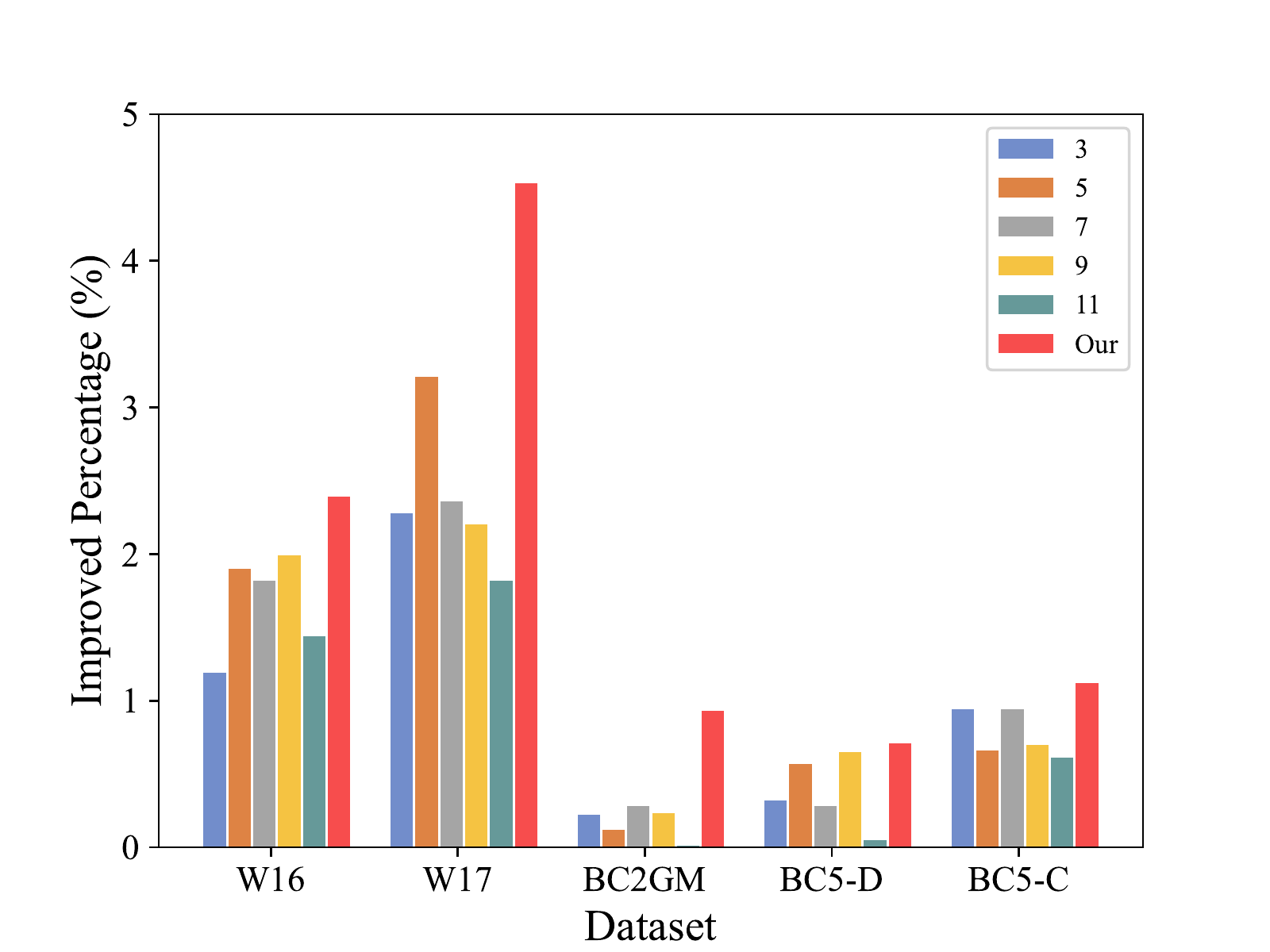}
\caption{\textcolor{black}{The improvement values (\%) of models with single windows or multi-window compared to Base models (BERT or BioBERT w.r.t. datasets), where 3, 5, 7, 9, 11 represents the single window size when models only use a single window to construct the Gang module.}}
\label{fig:ablation}
 \vskip -1em
\end{figure}

\paragraph{Effect of position information}
\textcolor{black}{
Recurrent structures are able to extract both context and position information by its token-by-token manner while other network structures, including CNN and MLP, fail to encode the relative position information.
Thus, to explore the effect of position information, we compare models with different structures to construct the Gang module and report the improvements of F-1 score based on different Gang modules in Figure \ref{fig:position}.
First, we can observe that models with Gang module are better than Base (i.e., BERT), where all the values in Figure \ref{fig:position} are positive, further illustrating the effectiveness of our innovation in combining both global and local features, no matter what type of structure is used to construct the Gang module.
Second, models with LSTM and GRU outperform those with CNN and MLP, indicating that recurrent structures are more promising in short sequence feature extraction.
Since the recurrent structures can effectively capture position information by its token-by-token manner and help the model understand word-word relations based on their relative positions, we may conclude that position information is vital for improving performance.
Third, the comparison between CNN and MLP shows the power of CNN in extracting features from sub-sequences since CNN can leverage more fine-grained features, such as n-gram.
}
\paragraph{Ablation studies}
In this subsection, we compare our multi-window model with single-window models, and the improvements compared with Base model are shown in Figure \ref{fig:ablation}.
We have following observations.
First of all, illustrated by the comparisons among Base (i.e., BERT) and others, models with sliding windows achieve better performance, where all the improvement values in Figure \ref{fig:ablation} are positive.
This illustrates that both single window and multi-window recurrent structures can help to enhance token representation and bring different degrees of improvement, which further shows the importance of local features in this task.
Second, we can observe that the optimal single window sizes for different datasets are also different.
For example, the optimal single window size of W17 is 5, while that for BC2GM is 7, which indicates that the best length of the local sequence depends on the characteristics of datasets to some extent.
Third, compared with those models using a single window, the multi-window recurrent module obtains better performance, illustrating that features extracted from multiple sub-sequences are more effective than those captured from a single one.
The reason could be that multi-window can help the model pay attention to different local context sub-sequences and give them appropriate weights through the multi-window attention mechanism, such that it can provide more reasonable local information and alleviate the impact of the characteristics of the datasets themselves.
\begin{figure*}[t]
\centering
\includegraphics[width=0.99\textwidth, trim=0 10 0 5]{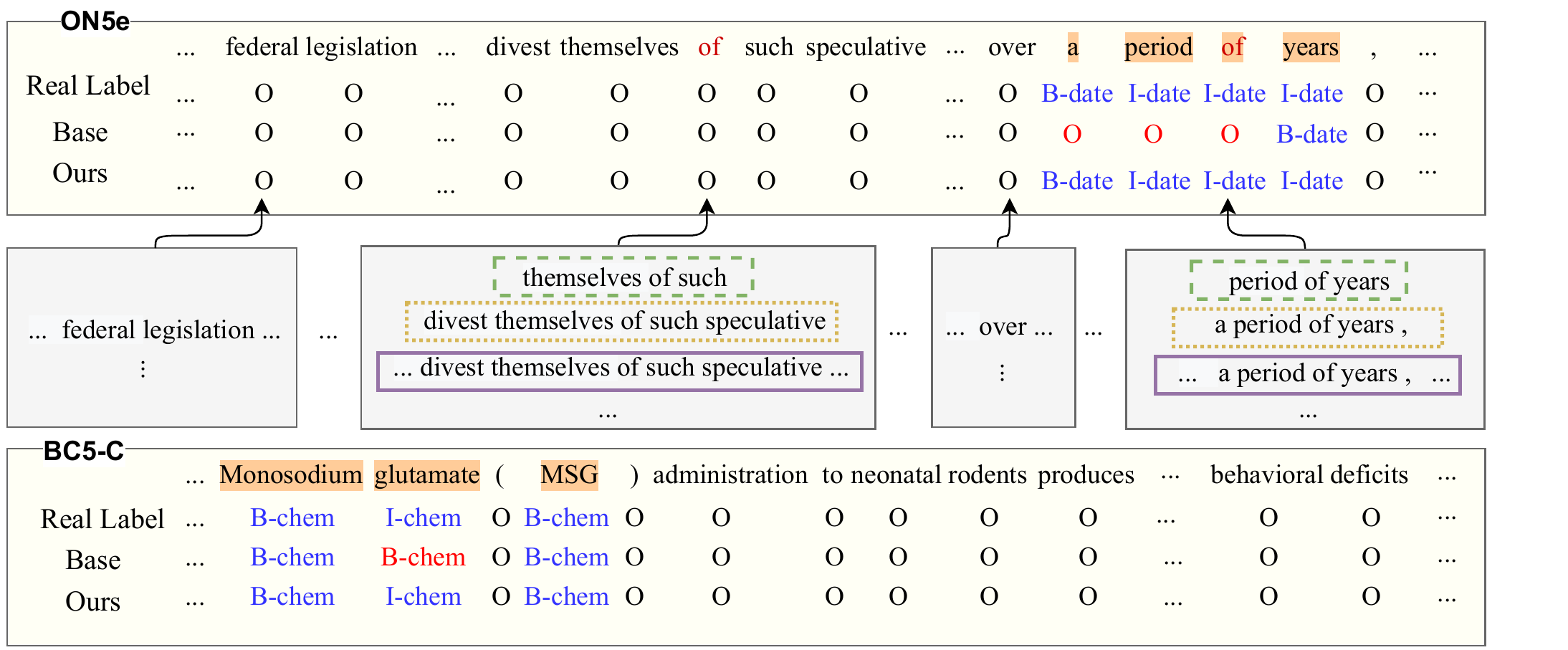}
\caption{\textcolor{black}{Examples of two predicted labels from \textsc{Base} and \textsc{Ours} as well as their corresponding source sentence and real label. Note that the \textsc{Base} for these two cases are \textsc{Bert} and \textsc{BioBERT}, respectively.}}
\label{fig:case}
 \vskip -1em
\end{figure*}

\paragraph{Case Study}
\textcolor{black}{
To further show the validity of our model, we perform qualitative analysis on some cases with their real labels and predicted labels from different models.
Figure \ref{fig:case} shows two cases from ON5e and BC5-C, respectively.
\textcolor{black}{We can observe that our model can predict more complete entities than Base. 
Specifically, in the first case, our model can recognize all the words in the entity \textit{"a period of years"} while Base model only recognizes the word \textit{"years"}.
In the second case, our model is able to identify \textit{"Monosodium glutamate"}, but Base model regards these words as two different entities.}
In addition, in the first example, compared with real labels, our model can label two \textit{"of"} correctly with the help of local features, which are O and I-date, respectively, while Base classifies both \textit{"of"} as O.
The sub-sequence (i.e.,\textit{"a period of years,"}) from the second \textit{"of"} is usually used to describe time such that this information is able to assist the model in marking the \textit{"of"} as I-date.
However, for the first \textit{"of"}, its sub-sequence \textit{"divest themselves of such speculative"} does not contain any meaning related to the entity themes, and thus the model marks the corresponding \textit{"of"} as O.
}

\section{Related Work}
\textcolor{black}{
NER is a fundamental task in NLP \cite{huang2015bidirectional}, which has drawn substantial attention over the past years and there have been many studies to address this task.
Recently, deep learning has played a dominant role in NER due to its effectiveness in capturing contextual information from sequences.
The recurrent neural networks (RNN), including its variants such as LSTM \cite{hochreiter1997long}, and GRU \cite{GRU}, is a promising structure for solving this task since it can effectively learn sequence information with its recurrent mechanism \cite{ma2016end, huang2015bidirectional, chiu2016named, zhu2019can}.
However, it is ineffective for RNN to learn long sequences due to the gradients exploding and vanishing.
Thus, Transformer-based models, such as BERT \cite{BERT}, BioBERT \cite{biobert}, and XLNET \cite{yangxlnet}, are proposed to alleviate these problems with the help of the self-attention mechanism.
Compared to RNN, Transformer is able to capture long-distance information through multiple multi-head attention layers and has achieved impressive performance in this task \cite{nie2020named,luo2020hierarchical,yamada2020luke, gui2019lexicon}.
}

However, multi-head attention usually treats every position identically, which lead to the loss of position information.
To mitigate this problem, several approaches have been proposed to advance the Transformer \cite{dai2019transformer,shaw2018self,yan2019tener}.
\citet{shaw2018self} proposed cross-lingual position representation to help self-attention alleviate word order divergences in different languages and learn position information.
\textcolor{black}{
\citet{yan2019tener} introduced the directional relative positional encoding and an adapted Transformer Encoder to model the character-level and word-level features.
}
Although these position embeddings are able to help the model learn position information, they are still not enough to solve the issue appropriately \cite{Rtransformer, huang2020improve,qu2021explore}.
Besides, Transformer-based approaches cannot effectively extract local features that are also important for sequence learning tasks, and some studies have been proposed to alleviate this problem \cite{xu2017local,li2019enhancing,yang2019convolutional}.
\citet{xu2017local} proposed to use the fixed-size ordinally forgetting encoding to model sentence fragments, which is then used to predict the label for each text fragment.
\citet{li2019enhancing} utilized convolutional self-attention by producing queries and keys with causal convolution to incorporate local contextual information into the attention mechanism.
To address these issues, we offer an alternative solution, namely Hero-Gang Neural model, to enhance local and position information extraction via multiple recurrent structures under the guidance of global information.

\section{Conclusion}
\textcolor{black}{
In this paper, we propose a novel Hero-Gang {N}eural (HGN) structure to effectively combine global and local features for enhancing NER.
In detail, the Hero module aims to capture global understanding by a Transformer-based encoder, which is then used to guide the Gang to extract local features and relative position information through a multi-window recurrent module.
Afterward, we utilize the multi-window attention to elaborately combine the global information and local features for enhancing representations that are then used to predict the entity label for each token.
Empirically, our proposed model achieves new state-of-the-art results on several NER benchmark datasets, including both general and biomedical domains.
Besides, we compare different structures to construct the Gang model and investigate the effect of the number of sliding windows, which further illustrates the effectiveness of our proposed model.
}

\section*{Acknowledgements}
This work is supported by Chinese Key-Area Research and Development Program of Guangdong Province (2020B0101350001), NSFC under the project “The Essential Algorithms and Technologies for Standardized Analytics of Clinical Texts” (12026610) and the Guangdong Provincial Key Laboratory of Big Data Computing, The Chinese University of Hong Kong, Shenzhen.

\bibliography{anthology,custom}

\begin{thebibliography}{47}
\expandafter\ifx\csname natexlab\endcsname\relax\def\natexlab#1{#1}\fi

\bibitem[{Beltagy et~al.(2019)Beltagy, Lo, and Cohan}]{beltagy2019scibert}
Iz~Beltagy, Kyle Lo, and Arman Cohan. 2019.
\newblock Scibert: {A} {P}retrained {L}anguage {M}odel for {S}cientific {T}ext.
\newblock In \emph{Proceedings of the 2019 Conference on Empirical Methods in
  Natural Language Processing and the 9th International Joint Conference on
  Natural Language Processing (EMNLP-IJCNLP)}, pages 3606--3611.

\bibitem[{Bosselut et~al.(2019)Bosselut, Rashkin, Sap, Malaviya, Celikyilmaz,
  and Choi}]{bosselut2019comet}
Antoine Bosselut, Hannah Rashkin, Maarten Sap, Chaitanya Malaviya, Asli
  Celikyilmaz, and Yejin Choi. 2019.
\newblock Comet: {C}ommonsense {T}ransformers for {A}utomatic {K}nowledge
  {G}raph {C}onstruction.
\newblock In \emph{Proceedings of the 57th Annual Meeting of the Association
  for Computational Linguistics}, pages 4762--4779.

\bibitem[{Chiu and Nichols(2016)}]{chiu2016named}
Jason~PC Chiu and Eric Nichols. 2016.
\newblock Named {E}ntity {R}ecognition with {B}idirectional {LSTM}-{CNN}s.
\newblock \emph{Transactions of the Association for Computational Linguistics},
  4:357--370.

\bibitem[{Cho et~al.(2014)Cho, van Merri{\"e}nboer, Gulcehre, Bahdanau,
  Bougares, Schwenk, and Bengio}]{GRU}
Kyunghyun Cho, Bart van Merri{\"e}nboer, Caglar Gulcehre, Dzmitry Bahdanau,
  Fethi Bougares, Holger Schwenk, and Yoshua Bengio. 2014.
\newblock Learning {P}hrase {R}epresentations using {RNN} {E}ncoder--{D}ecoder
  for {S}tatistical {M}achine {T}ranslation.
\newblock In \emph{Proceedings of the 2014 Conference on Empirical Methods in
  Natural Language Processing (EMNLP)}, pages 1724--1734.

\bibitem[{Dai et~al.(2019)Dai, Yang, Yang, Carbonell, Le, and
  Salakhutdinov}]{dai2019transformer}
Zihang Dai, Zhilin Yang, Yiming Yang, Jaime~G Carbonell, Quoc Le, and Ruslan
  Salakhutdinov. 2019.
\newblock Transformer-xl: {A}ttentive {L}anguage {M}odels beyond a
  {F}ixed-length {C}ontext.
\newblock In \emph{Proceedings of the 57th Annual Meeting of the Association
  for Computational Linguistics}, pages 2978--2988.

\bibitem[{Derczynski et~al.(2017)Derczynski, Nichols, van Erp, and
  Limsopatham}]{w17}
Leon Derczynski, Eric Nichols, Marieke van Erp, and Nut Limsopatham. 2017.
\newblock Results of the {WNUT}2017 {S}hared {T}ask on {N}ovel and {E}merging
  {E}ntity {R}ecognition.
\newblock In \emph{Proceedings of the 3rd Workshop on Noisy User-generated
  Text}, pages 140--147.

\bibitem[{Devlin et~al.(2018)Devlin, Chang, Lee, and Toutanova}]{BERT}
Jacob Devlin, Ming-Wei Chang, Kenton Lee, and Kristina Toutanova. 2018.
\newblock Bert: {P}re-training of {D}eep {B}idirectional {T}ransformers for
  {L}anguage {U}nderstanding.
\newblock \emph{arXiv preprint arXiv:1810.04805}.

\bibitem[{Gui et~al.(2019)Gui, Zou, Zhang, Peng, Fu, Wei, and
  Huang}]{gui2019lexicon}
Tao Gui, Yicheng Zou, Qi~Zhang, Minlong Peng, Jinlan Fu, Zhongyu Wei, and
  Xuan-Jing Huang. 2019.
\newblock A lexicon-based graph neural network for chinese ner.
\newblock In \emph{Proceedings of the 2019 Conference on Empirical Methods in
  Natural Language Processing and the 9th International Joint Conference on
  Natural Language Processing (EMNLP-IJCNLP)}, pages 1040--1050.

\bibitem[{He et~al.(2019)He, Guan, and Dai}]{he2019classifying}
Bin He, Yi~Guan, and Rui Dai. 2019.
\newblock Classifying {M}edical {R}elations in {C}linical {T}ext via
  {C}onvolutional {N}eural {N}etworks.
\newblock \emph{Artificial intelligence in medicine}, 93:43--49.

\bibitem[{Hochreiter et~al.(1997)Hochreiter, urgen Schmidhuber, and
  Elvezia}]{hochreiter1997long}
Sepp Hochreiter, J~urgen Schmidhuber, and Corso Elvezia. 1997.
\newblock Long {S}hort-term {M}emory.
\newblock \emph{Neural Computation}, 9(8):1735--1780.

\bibitem[{Huang et~al.(2021)Huang, Deng, Li, and Yuan}]{huang2021missformer}
Xiaohong Huang, Zhifang Deng, Dandan Li, and Xueguang Yuan. 2021.
\newblock {MISSF}ormer: {A}n {E}ffective {M}edical {I}mage {S}egmentation
  {T}ransformer.
\newblock \emph{arXiv preprint arXiv:2109.07162}.

\bibitem[{Huang et~al.(2020)Huang, Liang, Xu, and Xiang}]{huang2020improve}
Zhiheng Huang, Davis Liang, Peng Xu, and Bing Xiang. 2020.
\newblock Improve {T}ransformer {M}odels with {B}etter {R}elative {P}osition
  {E}mbeddings.
\newblock In \emph{Proceedings of the 2020 Conference on Empirical Methods in
  Natural Language Processing: Findings}, pages 3327--3335.

\bibitem[{Huang et~al.(2015)Huang, Xu, and Yu}]{huang2015bidirectional}
Zhiheng Huang, Wei Xu, and Kai Yu. 2015.
\newblock Bidirectional {LSTM}-{CRF} {M}odels for {S}equence {T}agging.
\newblock \emph{arXiv preprint arXiv:1508.01991}.

\bibitem[{Kingma and Ba(2014)}]{adam}
Diederik~P Kingma and Jimmy Ba. 2014.
\newblock Adam: A {M}ethod for {S}tochastic {O}ptimization.
\newblock \emph{arXiv preprint arXiv:1412.6980}.

\bibitem[{Lample et~al.(2016)Lample, Ballesteros, Subramanian, Kawakami, and
  Dyer}]{lample2016neural}
Guillaume Lample, Miguel Ballesteros, Sandeep Subramanian, Kazuya Kawakami, and
  Chris Dyer. 2016.
\newblock Neural {A}rchitectures for {N}amed {E}ntity {R}ecognition.
\newblock In \emph{Proceedings of the 2016 Conference of the North American
  Chapter of the Association for Computational Linguistics: Human Language
  Technologies}, pages 260--270.

\bibitem[{Lee et~al.(2020)Lee, Yoon, Kim, Kim, Kim, So, and Kang}]{biobert}
J~Lee, W~Yoon, S~Kim, D~Kim, S~Kim, CH~So, and J~Kang. 2020.
\newblock Biobert: a {P}re-trained {B}iomedical {L}anguage {R}epresentation
  {M}odel for {B}iomedical {T}ext {M}ining.
\newblock \emph{Bioinformatics (Oxford, England)}, 36(4):1234.

\bibitem[{Li et~al.(2019)Li, Jin, Xuan, Zhou, Chen, Wang, and
  Yan}]{li2019enhancing}
Shiyang Li, Xiaoyong Jin, Yao Xuan, Xiyou Zhou, Wenhu Chen, Yu-Xiang Wang, and
  Xifeng Yan. 2019.
\newblock Enhancing the {L}ocality and {B}reaking the {M}emory {B}ottleneck of
  {T}ransformer on {T}ime {S}eries {F}orecasting.
\newblock \emph{Advances in Neural Information Processing Systems},
  32:5243--5253.

\bibitem[{Liu et~al.(2019)Liu, Ott, Goyal, Du, Joshi, Chen, Levy, Lewis,
  Zettlemoyer, and Stoyanov}]{liu2019roberta}
Yinhan Liu, Myle Ott, Naman Goyal, Jingfei Du, Mandar Joshi, Danqi Chen, Omer
  Levy, Mike Lewis, Luke Zettlemoyer, and Veselin Stoyanov. 2019.
\newblock Roberta: {A} {R}obustly {O}ptimized {BERT} {P}retraining {A}pproach.
\newblock \emph{arXiv preprint arXiv:1907.11692}.

\bibitem[{Luo et~al.(2020)Luo, Xiao, and Zhao}]{luo2020hierarchical}
Ying Luo, Fengshun Xiao, and Hai Zhao. 2020.
\newblock Hierarchical {C}ontextualized {R}epresentation for {N}amed {E}ntity
  {R}ecognition.
\newblock In \emph{Proceedings of the AAAI Conference on Artificial
  Intelligence}, volume~34, pages 8441--8448.

\bibitem[{Ma and Hovy(2016)}]{ma2016end}
Xuezhe Ma and Eduard Hovy. 2016.
\newblock End-to-end {S}equence {L}abeling via {B}i-directional
  {LSTM}-{CNN}s-{CRF}.
\newblock In \emph{Proceedings of the 54th Annual Meeting of the Association
  for Computational Linguistics (Volume 1: Long Papers)}, pages 1064--1074.

\bibitem[{Morwal et~al.(2012)Morwal, Jahan, and Chopra}]{morwal2012named}
Sudha Morwal, Nusrat Jahan, and Deepti Chopra. 2012.
\newblock {N}amed {E}ntity {R}ecognition {U}sing {H}idden {M}arkov {M}odel
  (hmm).
\newblock \emph{International Journal on Natural Language Computing (IJNLC)
  Vol}, 1.

\bibitem[{Mozharova and Loukachevitch(2016)}]{mozharova2016combining}
Valerie~A Mozharova and Natalia~V Loukachevitch. 2016.
\newblock Combining {K}nowledge and {CRF}-based {A}pproach to {N}amed {E}ntity
  {R}ecognition in {R}ussian.
\newblock In \emph{International Conference on Analysis of Images, Social
  Networks and Texts}, pages 185--195. Springer.

\bibitem[{Nie et~al.(2020{\natexlab{a}})Nie, Tian, Song, Ao, and Wan}]{AESUBER}
Yuyang Nie, Yuanhe Tian, Yan Song, Xiang Ao, and Xiang Wan. 2020{\natexlab{a}}.
\newblock Improving {N}amed {E}ntity {R}ecognition with {A}ttentive {E}nsemble
  of {S}yntactic {I}nformation.
\newblock In \emph{Proceedings of the 2020 Conference on Empirical Methods in
  Natural Language Processing: Findings}, pages 4231--4245.

\bibitem[{Nie et~al.(2020{\natexlab{b}})Nie, Tian, Wan, Song, and
  Dai}]{nie2020named}
Yuyang Nie, Yuanhe Tian, Xiang Wan, Yan Song, and Bo~Dai. 2020{\natexlab{b}}.
\newblock Named {E}ntity {R}ecognition for {S}ocial {M}edia {T}exts with
  {S}emantic {A}ugmentation.
\newblock In \emph{Proceedings of the 2020 Conference on Empirical Methods in
  Natural Language Processing (EMNLP)}, pages 1383--1391.

\bibitem[{Peng et~al.(2019)Peng, Yan, and Lu}]{peng2019transfer}
Yifan Peng, Shankai Yan, and Zhiyong Lu. 2019.
\newblock Transfer {L}earning in {B}iomedical {N}atural {L}anguage
  {P}rocessing: An {E}valuation of {BERT} and {ELM}o on {T}en {B}enchmarking
  {D}atasets.
\newblock In \emph{Proceedings of the 18th BioNLP Workshop and Shared Task},
  pages 58--65.

\bibitem[{Pergola et~al.(2021)Pergola, Kochkina, Gui, Liakata, and
  He}]{pergola2021boosting}
Gabriele Pergola, Elena Kochkina, Lin Gui, Maria Liakata, and Yulan He. 2021.
\newblock Boosting {L}ow-{R}esource {B}iomedical {QA} via {E}ntity-{A}ware
  {M}asking {S}trategies.
\newblock In \emph{Proceedings of the 16th Conference of the European Chapter
  of the Association for Computational Linguistics: Main Volume}, pages
  1977--1985.

\bibitem[{Pradhan et~al.(2013)Pradhan, Moschitti, Xue, Ng, Bj{\"o}rkelund,
  Uryupina, Zhang, and Zhong}]{pradhan2013towards}
Sameer Pradhan, Alessandro Moschitti, Nianwen Xue, Hwee~Tou Ng, Anders
  Bj{\"o}rkelund, Olga Uryupina, Yuchen Zhang, and Zhi Zhong. 2013.
\newblock Towards {R}obust {L}inguistic {A}nalysis {U}sing {O}ntonotes.
\newblock In \emph{Proceedings of the Seventeenth Conference on Computational
  Natural Language Learning}, pages 143--152.

\bibitem[{Qu et~al.(2021)Qu, Niu, and Mo}]{qu2021explore}
Anlin Qu, Jianwei Niu, and Shasha Mo. 2021.
\newblock Explore {B}etter {R}elative {P}osition {E}mbeddings from {E}ncoding
  {P}erspective for {T}ransformer {M}odels.
\newblock In \emph{Proceedings of the 2021 Conference on Empirical Methods in
  Natural Language Processing}, pages 2989--2997.

\bibitem[{Sachan et~al.(2018)Sachan, Xie, Sachan, and
  Xing}]{sachan2018effective}
Devendra~Singh Sachan, Pengtao Xie, Mrinmaya Sachan, and Eric~P Xing. 2018.
\newblock Effective {U}se of {B}idirectional {L}anguage {M}odeling for
  {T}ransfer {L}earning in {B}iomedical {N}amed {E}ntity {R}ecognition.
\newblock In \emph{Machine learning for healthcare conference}, pages 383--402.
  PMLR.

\bibitem[{Shaw et~al.(2018)Shaw, Uszkoreit, and Vaswani}]{shaw2018self}
Peter Shaw, Jakob Uszkoreit, and Ashish Vaswani. 2018.
\newblock Self-attention with {R}elative {P}osition {R}epresentations.
\newblock In \emph{Proceedings of the 2018 Conference of the North American
  Chapter of the Association for Computational Linguistics: Human Language
  Technologies, Volume 2 (Short Papers)}, pages 464--468.

\bibitem[{Strauss et~al.(2016)Strauss, Toma, Ritter, De~Marneffe, and Xu}]{w16}
Benjamin Strauss, Bethany Toma, Alan Ritter, Marie-Catherine De~Marneffe, and
  Wei Xu. 2016.
\newblock Results of the {WNUT}16 {N}amed {E}ntity {R}ecognition {S}hared
  {T}ask.
\newblock In \emph{Proceedings of the 2nd Workshop on Noisy User-generated Text
  (WNUT)}, pages 138--144.

\bibitem[{Tong et~al.(2021)Tong, Chen, and Shi}]{2021A}
Y.~Tong, Y.~Chen, and X.~Shi. 2021.
\newblock A {M}ulti-{T}ask {A}pproach for {I}mproving {B}iomedical {N}amed
  {E}ntity recognition by incorporating multi-granularity information.
\newblock In \emph{Findings of the Association for Computational Linguistics:
  ACL-IJCNLP 2021}.

\bibitem[{Vaswani et~al.(2017)Vaswani, Shazeer, Parmar, Uszkoreit, Jones,
  Gomez, Kaiser, and Polosukhin}]{vaswani2017attention}
Ashish Vaswani, Noam Shazeer, Niki Parmar, Jakob Uszkoreit, Llion Jones,
  Aidan~N Gomez, {\L}ukasz Kaiser, and Illia Polosukhin. 2017.
\newblock Attention is {A}ll you {N}eed.
\newblock In \emph{Proceedings of the 31st International Conference on Neural
  Information Processing Systems}, pages 6000--6010.

\bibitem[{Wang et~al.(2020)Wang, Xu, Fu, Xu, and Wu}]{wang2020astral}
Jiuniu Wang, Wenjia Xu, Xingyu Fu, Guangluan Xu, and Yirong Wu. 2020.
\newblock {ASTRAL}: {A}dversarial {T}rained {LSTM}-{CNN} for {N}amed {E}ntity
  {R}ecognition.
\newblock \emph{Knowledge-Based Systems}, 197:105842.

\bibitem[{Wang et~al.(2021)Wang, Jiang, Bach, Wang, Huang, Huang, and
  Tu}]{wang2021improving}
Xinyu Wang, Yong Jiang, Nguyen Bach, Tao Wang, Zhongqiang Huang, Fei Huang, and
  Kewei Tu. 2021.
\newblock Improving {N}amed {E}ntity {R}ecognition by {E}xternal {C}ontext
  {R}etrieving and {C}ooperative {L}earning.
\newblock \emph{arXiv preprint arXiv:2105.03654}.

\bibitem[{Wang et~al.(2019{\natexlab{a}})Wang, Zhang, Ren, Zhang, Zitnik,
  Shang, Langlotz, and Han}]{wang2019cross}
Xuan Wang, Yu~Zhang, Xiang Ren, Yuhao Zhang, Marinka Zitnik, Jingbo Shang,
  Curtis Langlotz, and Jiawei Han. 2019{\natexlab{a}}.
\newblock Cross-type {B}iomedical {N}amed {E}ntity {R}ecognition with {D}eep
  {M}ulti-task {L}earning.
\newblock \emph{Bioinformatics}, 35(10):1745--1752.

\bibitem[{Wang et~al.(2019{\natexlab{b}})Wang, Ma, Liu, and
  Tang}]{Rtransformer}
Zhiwei Wang, Yao Ma, Zitao Liu, and Jiliang Tang. 2019{\natexlab{b}}.
\newblock R-transformer: {R}ecurrent {N}eural {N}etwork {E}nhanced
  {T}ransformer.
\newblock \emph{arXiv preprint arXiv:1907.05572}.

\bibitem[{Wolf et~al.(2020)Wolf, Chaumond, Debut, Sanh, Delangue, Moi, Cistac,
  Funtowicz, Davison, Shleifer et~al.}]{wolf2020transformers}
Thomas Wolf, Julien Chaumond, Lysandre Debut, Victor Sanh, Clement Delangue,
  Anthony Moi, Pierric Cistac, Morgan Funtowicz, Joe Davison, Sam Shleifer,
  et~al. 2020.
\newblock Transformers: {S}tate-of-the-art {N}atural {L}anguage {P}rocessing.
\newblock In \emph{Proceedings of the 2020 Conference on Empirical Methods in
  Natural Language Processing: System Demonstrations}, pages 38--45.

\bibitem[{Xu et~al.(2021)Xu, Jie, Lu, and Bing}]{xu2021better}
Lu~Xu, Zhanming Jie, Wei Lu, and Lidong Bing. 2021.
\newblock Better {F}eature {I}ntegration for {N}amed {E}ntity {R}ecognition.
\newblock In \emph{Proceedings of the 2021 Conference of the North American
  Chapter of the Association for Computational Linguistics: Human Language
  Technologies}, pages 3457--3469.

\bibitem[{Xu et~al.(2017)Xu, Jiang, and Watcharawittayakul}]{xu2017local}
Mingbin Xu, Hui Jiang, and Sedtawut Watcharawittayakul. 2017.
\newblock A {L}ocal {D}etection {A}pproach for {N}amed {E}ntity {R}ecognition
  and {M}ention {D}etection.
\newblock In \emph{Proceedings of the 55th Annual Meeting of the Association
  for Computational Linguistics (Volume 1: Long Papers)}, pages 1237--1247.

\bibitem[{Yamada et~al.(2020)Yamada, Asai, Shindo, Takeda, and
  Matsumoto}]{yamada2020luke}
Ikuya Yamada, Akari Asai, Hiroyuki Shindo, Hideaki Takeda, and Yuji Matsumoto.
  2020.
\newblock {LUKE}: {D}eep {C}ontextualized {E}ntity {R}epresentations with
  {E}ntity-{A}ware {S}elf-{A}ttention.
\newblock In \emph{Proceedings of the 2020 Conference on Empirical Methods in
  Natural Language Processing (EMNLP)}, pages 6442--6454.

\bibitem[{Yan et~al.(2019)Yan, Deng, Li, and Qiu}]{yan2019tener}
Hang Yan, Bocao Deng, Xiaonan Li, and Xipeng Qiu. 2019.
\newblock Tener: {A}dapting {T}ransformer {E}ncoder for {N}amed {E}ntity
  {R}ecognition.
\newblock \emph{arXiv preprint arXiv:1911.04474}.

\bibitem[{Yan et~al.(2021)Yan, Gui, Dai, Guo, Zhang, and Qiu}]{yan2021unified}
Hang Yan, Tao Gui, Junqi Dai, Qipeng Guo, Zheng Zhang, and Xipeng Qiu. 2021.
\newblock A unified generative framework for various ner subtasks.
\newblock In \emph{Proceedings of the 59th Annual Meeting of the Association
  for Computational Linguistics and the 11th International Joint Conference on
  Natural Language Processing (Volume 1: Long Papers)}, pages 5808--5822.

\bibitem[{Yang et~al.(2019{\natexlab{a}})Yang, Wang, Wong, Chao, and
  Tu}]{yang2019convolutional}
Baosong Yang, Longyue Wang, Derek~F Wong, Lidia~S Chao, and Zhaopeng Tu.
  2019{\natexlab{a}}.
\newblock Convolutional {S}elf-{A}ttention {N}etworks.
\newblock In \emph{Proceedings of the 2019 Conference of the North American
  Chapter of the Association for Computational Linguistics: Human Language
  Technologies, Volume 1 (Long and Short Papers)}, pages 4040--4045.

\bibitem[{Yang et~al.(2019{\natexlab{b}})Yang, Dai, Yang, Carbonell,
  Salakhutdinov, and Le}]{yangxlnet}
Zhilin Yang, Zihang Dai, Yiming Yang, Jaime Carbonell, Russ~R Salakhutdinov,
  and Quoc~V Le. 2019{\natexlab{b}}.
\newblock Xlnet: {G}eneralized {A}utoregressive {P}retraining for {L}anguage
  {U}nderstanding.
\newblock \emph{Advances in Neural Information Processing Systems},
  32:5753--5763.

\bibitem[{Yuan et~al.(2021)Yuan, Liu, Tan, Huang, and
  Huang}]{yuan2021improving}
Zheng Yuan, Yijia Liu, Chuanqi Tan, Songfang Huang, and Fei Huang. 2021.
\newblock Improving {B}iomedical {P}retrained {L}anguage {M}odels with
  {K}nowledge.
\newblock In \emph{Proceedings of the 20th Workshop on Biomedical Language
  Processing}, pages 180--190.

\bibitem[{Zhu and Wang(2019)}]{zhu2019can}
Yuying Zhu and Guoxin Wang. 2019.
\newblock Can-ner: Convolutional attention network for chinese named entity
  recognition.
\newblock In \emph{Proceedings of the 2019 Conference of the North American
  Chapter of the Association for Computational Linguistics: Human Language
  Technologies, Volume 1 (Long and Short Papers)}, pages 3384--3393.

\end{thebibliography}
\bibliographystyle{acl_natbib}

\clearpage
\appendix
\section{Appendix}
\label{sec:appendix}

\subsection{Hyper-parameter Settings}
\label{appendix:hyperparameter}
\begin{table*}[t]
\centering
\vspace{1mm}
\begin{tabular}{c|c|c|c|c|c|c|c}
\toprule
\multirow{2}{*}{\textsc{\textbf{Model}}} & \multirow{2}{*}{\textsc{\textbf{Hypter.}}} & \multicolumn{3}{c|}{\textsc{\textbf{NLP Data}}} & \multicolumn{3}{c}{\textsc{\textbf{Biomedicine Data}}} \\ \cmidrule(l){3-8} 
 &  & \textsc{W16} & \textsc{W17} & \textsc{ON5e} & \textsc{BC2GM} & \textsc{BC5-D} & \textsc{BC5-C} \\ \midrule
\multirow{3}{*}{\textsc{HGN (MLP)}} & Window Size &\{1,3,5,7\} &  \{3,5,7\}
&  \{5,7,9\}
&  \{1,3,5\}
&  \{5,7,11\}
&  \{5,7,11\}
\\
 & Learning Rate & 3e-5 & 5e-5 & 1e-5 & 1e-5 & 9e-6 &1e-5  \\
 & Batch Size & 32  & 32 & 32 & 32 & 32 &32  \\ \midrule
\multirow{3}{*}{\textsc{HGM (DOT)}} & Window Size & \{3,5,7\} & \{5,7,9\}
&  \{3,5,7\}
&  \{3,5,7\}
&  \{5,7,9\}
&  \{5,7,11\}
\\
 & Learning Rate &3e-5  &5e-5  & 1e-5 & 1e-5 &9e-6  &9e-6  \\
 & Batch Size & 32  & 32 & 32 & 32 & 32 &32  \\ \bottomrule
\end{tabular}
\caption{The hyper-parameters for best models that we have experimented on the given datasets.}
\label{tab:hypter-parameter}
\end{table*}
We have tested several combinations of hyper-parameters in tuning our models for all NLP and Biomedical benchmark datasets (i.e., W16, W17, \textsc{ON5e}, BC5CDR-disease, BC2GM, and BC5CDR-chem). Table \ref{tab:hypter-parameter} reports the combinations that achieve the highest F-1 score for each dataset.

\end{document}